\UseRawInputEncoding
\documentclass[review,12pt,a4papper]{elsarticle}
\usepackage{amsmath,amssymb}

\usepackage{multirow}
\usepackage{float}
\usepackage{makecell}
\usepackage{algorithm}
\usepackage{algorithmic}
\usepackage{graphicx}
\usepackage{bm}
\usepackage{subfigure}
\usepackage{booktabs}
\usepackage{array, caption, threeparttable}
\usepackage{appendix}
\usepackage[section]{placeins}
\usepackage[font=small,labelfont=bf]{caption}
\usepackage{lineno,hyperref}
\modulolinenumbers[5]

\begin{document}

\begin{frontmatter}

\title{Multi-view Feature Extraction based on Dual Contrastive Head}

\author{Hongjie Zhang}
\address{College of Information and Electrical Engineering, China Agricultural University, Beijing 100083, China}

\begin{abstract}
	Multi-view feature extraction is an efficient approach for alleviating the issue of dimensionality in highdimensional
	multi-view data. Contrastive learning (CL), which is a popular self-supervised learning method, has
	recently attracted considerable attention. Most CL-based methods were constructed only from the sample level. In this study, we propose
	a novel multi-view feature extraction method based on dual contrastive head, which introduce structural-level contrastive loss into sample-level CL-based method. Structural-level CL push the potential subspace structures consistent in any two cross views, which assists sample-level CL to extract discriminative features more effectively. Furthermore, it is proven that the relationships between structural-level CL and mutual information and probabilistic intra- and inter- scatter, which provides the theoretical support for the excellent performance.
	Finally, numerical experiments on six real datasets demonstrate the superior performance of the proposed method compared to existing methods.
\end{abstract}


\end{frontmatter}


\section{Introduction}
At present, multi-view data has been widely used in machine learning tasks such as classification and clustering\cite{c1,cluster1,cluster2,e1,s1}, as it contains richer feature information compared to single-view data. However, multi-view data have often appeared in a high dimensional form, which results in a considerable
waste of time and costs as well as causes the problem known as the “curse of dimensionality”.  An effective means of addressing these issues is multi-view feature extraction, which transforms the original high-dimensional spatial data into a low-dimensional
subspace using some projection matrices\cite{ex1,ex2,ex3}. Although the effect
of feature extraction is often slightly worse than that of deep
learning, it has always been a research hotspot owing to its
strong interpretability and compatibility with any type of hardware
(CPU, GPU, and DSP). Therefore, there is an urgent need for
traditional multi-view feature extraction methods to extract discriminative
features for downstream tasks more effectively. \par 
Contrastive learning (CL)\cite{cl1,cl2,cl3,cl4,cl5}, as the primary method of self-supervised learning, has become a popular topic for processing multi-view data in the field of deep learning. InfoNCE loss based on CL is proposed in contrastive predictive coding\cite{cpc} and has demonstrated excellent performance. Consequently, a large number of studies based on CL are proposed. Tian et al. proposed contrastive multiview coding (CMC)\cite{cmc} to process multi-view data. CMC first constructs the same sample in any two views as positive pairs and distinct samples as negative pairs, and subsequently, optimizes a neural network by minimizing the InfoNCE loss to maximize the similarity of the projected positive pairs. Chen et al. proposed a simple framework for CL (SimCLR)\cite{simclr} to obtain the representations of multiple views by performed data augmentation for single-view data, following which it optimizes the network by minimizing the InfoNCE loss. Due to the number of negative pairs directly affects the performance of CL, a number of contrastive approaches either substantially increase
the batch size, or keep large memory banks. For example, non-parametric softmax classifier\cite{npid} and pretext-invariant representation learning\cite{pirl} use memories that contain the whole training set, while the momentum contrast\cite{moco} of He et al. keeps a queue with features of the last few batches as memory. It is however shown that increasing the memory or batch size leads to diminishing returns in terms of performance, suggesting that more negative pairs does not necessarily mean hard negative pairs. Therefore, invariance propagation (InvP)\cite{invp} and mixing of contrastive hard negatives (MoCHi)\cite{mochi} based on distinct hard sampling strategies were proposed. InvP focus on learning representations invariant to category-level
variations by optimizing a network, which are provided by structural information of the original samples. MoCHi constructs the hardest negative pairs and even harder negative pairs in the original space, and subsequently, optimizes a neural network by minimizing the InfoNCE loss. Prototypical contrastive learning (PCL)\cite{pcl} and hiearchical contrastive selective coding (HCSC)\cite{hcsc}, which bridges CL with clustering. PCL introduces prototypes as latent variables to help find the maximum-likelihood estimation of the network parameters in an Expectation-Maximization framework, while HCSC propose to represent the hierarchical semantic structures of image representations by dynamically maintaining hierarchical prototypes.\par 
Although the above CL-based methods have good performance, they also have some disadvantages. Firstly, the existing methods are produced in the field of deep learning and are not suitable for obtaining projection matrices to process the traditional multi-view feature extraction problem. Secondly, the methods based on hard sampling strategies pre-define positive and negative samples in original space, which takes the risk of choosing pseudo-positive and -negative pairs.  Furthermore, the methods based on clustering are prone to false labels when the number of clusters is not equal to the number of true categories, which will lead to the dispersion of embedded
samples from the same class and the aggregation of embedded
samples from distinct classes.\par 
Inspired by previous studies, we propose a novel multi-view feature extraction method based on dual contrastive head (MFEDCH), which introduces structural-level contrastive loss into the sample-level CL-based method. Concretely, we construct the structural-level contrastive loss from the potential structural information based on the result of feature extraction, which defines the adaptive self-reconstruction coefficients of the same embedded sample as positive pairs and the adaptive self-reconstruction coefficients of distinct embedded samples as negative pairs in any cross views. In particular, structural-level CL push the potential subspace structures consistent in any cross views, which assists sample-level CL to extract discriminative features more effectively. 
In fact, there are two potential tasks in MFEDCH, which are to obtain the projection matrices and to explore the subspace structures. By leveraging the interactions between these two essential tasks, MFEDCH are able to explore more real and reliable subspace structures, and obtain the optimal projection matrices. 
Moreover, MFEDCH avoids the impact of false labels due to errors of pre-determined the number of clusters.
In addition, in order to make our framework more generalizable, the balance parameter $\lambda$ is introduced into MFEDCH to balance the importance of sample- and structural-level contrastive loss.\par 
The main contributions of this study are as follows:
\begin{itemize}
	\item A novel unsupervised multi-view feature extraction method based on dual contrastive head (MFEDCH) is proposed, which introduces structural-level contrastive loss into the sample-level CL-based method.
	\item The structural-level contrastive loss is constructed from the potential structural information, which push the real and reliable subspace structures consistent in any cross views.
	\item It is proven that the relationships between structural-level CL and mutual information and probabilistic intra- and inter- scatter, which provides the theoretical support for the excellent performance.
	\item The experiments on six real-word datasets show the advantages of the proposed method.
\end{itemize}
\par 
The remainder of this article is organized as follows. Traditional multi-view  feature extraction methods are briefly introduced in Section \uppercase\expandafter{\romannumeral2}. Subsequently, the preliminary study and analysis are discussed in Section \uppercase\expandafter{\romannumeral3}. Extensive experiments that were conducted on several real-world datasets are outlined in Section \uppercase\expandafter{\romannumeral4}. Finally, the conclusion of this study are presented in Section \uppercase\expandafter{\romannumeral5}.

\section{Related Work}
Several multi-view feature extraction methods have been proposed in recent years, which can be categorized as unsupervised, semi-supervised, and supervised methods. In this study, we focus on unsupervised feature
extraction. The most classical and representative method is canonical correlation analysis (CCA)\cite{cca}, which finds two projection matrices by maximizing the correlation of distinct representations of the same embedded sample of two views. However, CCA only suitable for linear scenario. In order to deal with the nonlinear scenario, kernel canonical correlation analysis is proposed which combines kernel trick with linear CCA. As the developments of mainfold learning, local preserving canonical correlation analysis (LPCCA)\cite{lpcca} and a new LPCCA (ALPCCA)\cite{alpcca} were proposed to preserve local structure of nonlinear two-view data. LPCCA finds projection matrices by preserving local neighbor information in each view, while ALPCCA explores extra cross-view correlations between neighbors. Nevertheless, the number of neighbors in
both methods is manually chosen by experience, which affects
the final results. Thus, Zu et al. proposed canonical sparse cross-view correlation analysis (CSCCA)\cite{cscca} which combines sparse
reconstruction and LPCCA to explore local intrinsic geometric
structure automatically. Zhu et al. pointed out that CSCCA neglects
the weights of data, as the difference among samples is not
well modeled. Therefore, a weight-based CSCCA (WCSCCA)\cite{wcscca}
was proposed. WCSCCA measures the correlation between two
views using the weights of data and the cross-view information. Zhao et al. proposed co-training
locality preserving projections\cite{colpp} which aims at finding
a low-dimensional embedding such that the local neighbor
structures of two views are maximumly compatible.\par 
However, all of the above methods are only suitable for
two-view data. In order to deal with multi-view data,
Foster et al. proposed Multi-view CCA\cite{mcca} which finds a common subspace by maximizing the total canonical correlation
coefficients between any two views. As a further extension, Cao et al. proposed multi-view partial least squares\cite{mvpls}. By unifying
Laplacian eigenmaps\cite{le} and multi-view learning, Xia et al.
proposed multi-view spectral embedding (MSE)\cite{mse}. MSE finds a
subspace in which the low-dimensional embedding is sufficiently
smooth. Wang et al. adopted a novel locality linear embedding
scheme to develop a new method named multi-view reconstructive
preserving embedding (MRPE)\cite{mrpe}. Combining sparse reconstruction
and co-regularized scheme, a co-regularized multi-view
sparse reconstruction embedding (CMSRE)\cite{cmsre} was proposed. A
common characteristic of MSE, MRPE and CMSRE is that they
find the low-dimensional embedding directly, and it can not obtain the embedding of a new
sample. Therefore, sparsity preserving multiple canonical correlation
analysis\cite{spmcca} and graph multi-view canonical
correlation analysis\cite{gmcca} were proposed. Wang et al. proposed kernelized multi-view subspace analysis\cite{kmsa} which can automatically learn the weights for all views and extract the nonlinear
features simultaneously. Zhang et al. proposed standard locality cross-view regression (SLCR)\cite{slcr}, which uses a regression loss to explore the reconstructed relationship in any cross views.\par

\section{Preliminary Study and Analysis}
\subsection{Notation and Definition}
This study considers the problem of unsupervised multi-view  feature extraction. Let us mathematically formulate this problem as follows.\par 
Multi-view feature extraction problem: Given $V$ training sample sets $X^m=[x_1^m,x_2^m,...,x_n^m]\in R^{D_m\times n}, m=1,2,...,V$ from $V$ different views, where $X^m , m=1,2,...,V$ represents the dataset from the $m$th view, $x_i^m, i=1,2,...,n$ represents the $i$th sample in the $m$th view. $D_m$ represents the feature dimension in the $m$th view. The purpose of feature extraction is to find projection matrices $P_m\in R^{D_m\times d}, m=1,2,...,V$ to derive the low-dimensional embeddings $Y^m=[y_1^m,y_2^m,...,y_n^m]\in R^{d\times n}$ for $X^m$ calculated by $P_m^TX^m$, where $d\ll D^m$.\par
\subsection{Preliminary Study}
\subsection*{\rm{\textbf{Sample-level CL}}}
The basic idea of sample-level CL is to optimize the neural network by maximizing the similarity of positive pairs ($x_i^m, x_{i}^{m +}$) and minimizing the similarity of negative pairs ($x_i^m, x_{i}^{m -}$) in the InfoNCE loss. The main difference between current CL-based methods is the strategy for constructing positive and negative pairs.  To perform traditional multi-view feature extraction, the InfoNCE loss is specified as follows:
\begin{small}
	\begin{equation}
	\begin{aligned}\label{equ1}
	&\mathcal L^\theta_{\text {InfoNCE}}(P_m,m=1,...,V)=\sum_{m=1}^{V}\mathbb{E}_{x^m_i, x_i^{m +}, x_{i}^{m-}}\\
	&\left[-\log \frac{\sum_{x_i^{m+}}\operatorname{exp}(\operatorname{sim}\left(y^m_i,  y_i^{m +}\right)}{\sum_{x_i^{m+}}\operatorname{exp}(\operatorname{sim}\left(y_i^m,  y_i^{m +}\right)+ \sum_{x_i^{m-}}\operatorname{exp}(\operatorname{sim}\left(y_i^m,  y_i^{m -}\right)}\right],\\
	\end{aligned}
	\end{equation}
\end{small}
where 
\begin{equation}
\operatorname{sim}\left(y_i^m, y_i^{m +}\right)=\frac{y_i^{m T} y_i^{m+}}{\left\|y_i^m\right\|\left\|y_i^{m +}\right\|\tau_1},
\end{equation}
$\theta$ can be selected from traditional CL-based methods, such as CMC, SimCLR, InvP, MoCHi, etc., representing the definition of positive and negative samples in a manner consistent with them, respectively. $\tau_1$ is a scalar temperature parameter.\par 
Although these CL-based methods mentioned in the previous section  have good performance in unsupervised learning, they also have some disadvantages. Firstly, the existing CL-based methods are produced in the field of deep learning and are not suitable for obtaining the projection matrices to process the traditional multi-view
feature extraction problem. Secondly, even if we construct
InfoNCE loss like (\ref{equ1}) according to these existing methods of defining
positive and negative pairs, and use it to perform feature
extraction, there suffers from the following problems: these algorithms like SimCLR, CMC, and MoCo only define different representations of the same samples as positive pairs and different samples as negative pairs, without considering the structural information of the data, which lead to the dispersion of embedded samples from the same class and the aggregation of embedded samples from distinct classes; moreover,
the definitions of positive and negative pairs like InvP and MoCHi consider structural information of original samples,  which runs the risk of pseudo-positive and -negative pairs due to unrealistic and unreliable structures being generated through noise and redundant features of the original samples; furthermore, the CL-based algorithms like PCL and HCSC are prone to false labels when the number of clusters is not equal to the number of true categories, which also lead to the dispersion of embedded samples from the same class and the aggregation of embedded samples from distinct classes.\par 
\subsection*{\rm{\textbf{Structural-level CL}}}

To solve the above problems, the structural-level CL is proposed, which construct the InfoNCE loss from the potential subspace structural information based on the result of feature extraction. Specifically, the adaptive self-reconstruction coefficients of the same embedded sample are defined as positive pairs and the adaptive self-reconstruction coefficients of distinct embedded samples are defined as negative pairs in any cross views. The structural-level contrastive loss is as follows:
\begin{equation}
\begin{aligned}
&\mathcal L_{\text {str}}(P_m,W^m,m=1,...,V)\\=&\sum_{m=1}^{V}\sum_{v=1,v\neq m}^{V}\mathbb{E}_{w_i^m, w_i^{v},\left\{w_{k}^{v}\right\}_{k=1}^{n}}\left[-\log \frac{\operatorname{exp}(\operatorname{sim}\left(w_i^m,  w_i^v\right)}{\sum_{k=1}^{n} \operatorname{exp}(\operatorname{sim}\left(w^m_i,  w_k^{v}\right)}\right]\\
&+\sum_{m=1}^{V}\left(\alpha\left\|Y^m-Y^mW^m\right\|_{F}^{2}+\beta\left\|W^m\right\|_F^2\right),
\end{aligned}
\end{equation}
where
\begin{equation}
\operatorname{sim}\left(w_i^m, w_k^v \right)=\frac{{w^m_i}^T w^v_k}{\left\|w^m_i\right\|\left\|w^v_k\right\|\tau_2},
\end{equation}
$\alpha$ and $\beta$ are two positive parameters, $\tau_2$ is a scalar temperature parameter, $w^m_i$ is the $i$th column vector of $W^m$. The first term of the loss function constrains the adaptive self-reconstruction coefficients of the same embedded samples in cross views to be consistent, the second term reconstructs adaptively the embedded samples, which obtains the potential structural information based on the result of feature extraction, the third term prevent $W^m$ from being a identity matrix.\par 
Structural-level CL incorporates the exploration of the potential structural information of representing accurate categories into the procedure of multi-view feature extraction through pushing the subspace structures consistent in any cross views. At the same time, the common real and reliable subspace structures are used within each view to ensure that the intra-class embedded samples are more aggregated and the inter-class embedded samples are more dispersed, so as to extract discriminative features more effectively. \par

\subsection*{\rm {\textbf{MFEDCH}}}
We propose a novel CL-based method MFEDCH, which construct contrastive loss by introducing structural-level contrastive loss into sample-level CL-based method CMC. The specific loss function is as follows:
\begin{equation}\label{MFEDCH}
\mathcal L(P_m,W^m,\\m=1,...,V)=\mathcal L^{\text{CMC}}_{\text {InfoNCE}}+\lambda \mathcal L_{\text {str}}
\end{equation}
where $\lambda$ is a positive parameter to balance the importance of sample- and structural-level contrastive loss.\par 
The proposed MFEDCH contains the sample- and structural-level CL,  which construct positive and negative pairs from subspace samples and potential structural information, respectively.
In our MFEDCH, structural-level CL push the potential subspace structures consistent in any cross views, which assists sample-level CL to extract discriminantive features more effectively. 
In fact, there are two potential tasks, which are to obtain the projection matrices and to explore the subspace structures. By leveraging the interactions between these two essential tasks, MFEDCH are able to explore more real and reliable subspace structures of representing accurate categories, and obtain the optimal projection matrices.
In addition, strucural-level CL can also be combined with other sample-level CL-based methods, such as SimCLR, InvP, MoCHi, PCL, HCSC, etc., which provides a new framework for future CL.\par 
\subsection{Theoretical Analysis}
\subsection*{\rm {\textbf{Structural-level CL and Mutual Information}}}
The relationship between structural-level CL and mutual information are analyzed from the theoretical aspect.
Obviously, when $m$ and $v (m\neq v)$ are fixed, $w^v_k$ is a positive sample of $w^m_i$ iff $i=k$, otherwise $w^v_k$ is a negative sample of $w^m_i$. Naturally, the probability that the sample $w^v_k$ in $W^v$ is a positive sample of $w^m_i$ is equal to $p(i=k |w^v_k, w^m_i)$, and the structural-level loss is equivalent to:
\begin{equation}
\begin{aligned}
\mathcal L_{\text{str}}=&\sum_{m=1}^{V}\sum_{v=1,v\neq m}^{V}\mathbb{E}_{w_i^m, w_k^{v}}\left[-\log   p(i=k|w^v_k, w^m_i)\right]\\
&+\sum_{m=1}^{V}\left(\alpha\left\|Y^m-Y^mW^m\right\|_{F}^{2}+\beta\left\|W^m\right\|_F^2\right).
\end{aligned}
\end{equation}
Therefore, there are two prior distributions $p(i=k) = \frac{1}{n}$ and $p(i\neq k) = \frac{n-1}{n}$. According to the Bayesian formula, the following derivation is made:
\begin{equation}
\begin{aligned}
&p(i=k|w^v_k, w^m_i)\\
&=\frac{p(w^v_k, w^m_i|i=k)p(i=k)}{p(w^v_k, w^m_i|i=k)p(i=k)+p(w^v_k, w^m_i|i\neq k)p(i\neq k)}\\
&=\frac{p(w^v_k, w^m_i|i=k)\frac{1}{n}}{p(w^v_k, w^m_i|i=k)\frac{1}{n}+p(w^v_k, w^m_i|i\neq k)\frac{n-1}{n}}\\
&=\frac{p(w^v_k, w^m_i|i=k)}{p(w^v_k, w^m_i|i=k)+(n-1)p(w^v_k, w^m_i|i\neq k)}\\
&=\frac{p(w^v_k, w^m_i)}{p(w^v_k, w^m_i)+(n-1)p(w^v_k)p(w^m_i)}.\\
\end{aligned}
\end{equation}
Further derivation, there is the following formula:
\begin{small}
	\begin{equation}
	\begin{aligned}
	&\mathcal L_{\text {str}}\\
	&\geq\sum_{m=1}^{V}\sum_{v=1,v\neq m}^{V}\mathbb{E}_{w_i^m, w_k^{v}}\left[-\log   p(i=k|w^v_k, w^m_i)\right]\\
	&=\sum_{m=1}^{V}\sum_{v=1,v\neq m}^{V}\mathbb{E}_{w_i^m, w_k^{v}}\left[-\log  \frac{p(w^v_k, w^m_i)}{p(w^v_k, w^m_i)+(n-1)p(w^v_k)p(w^m_i)}\right]\\
	&=\sum_{m=1}^{V}\sum_{v=1,v\neq m}^{V}\mathbb{E}_{w_i^m, w_k^{v}}\left[\log  \frac{p(w^v_k, w^m_i)+(n-1)p(w^v_k)p(w^m_i)}{p(w^v_k, w^m_i)}\right]\\
	&=\sum_{m=1}^{V}\sum_{v=1,v\neq m}^{V}\mathbb{E}_{w_i^m, w_k^{v}}\log  \left[1+\left( n-1\right) \frac{p(w^v_k)p(w^m_i)}{p(w^v_k, w^m_i)}\right]\\
	&=\sum_{m=1}^{V}\sum_{v=1,v\neq m}^{V}\mathbb{E}_{w_i^m, w_k^{v}}\log \left[ \frac{p(w^v_k,w^m_i)-p(w^v_k)p(w^m_i)}{p(w^v_k,w^m_i)}\right.\\
	&\quad\left.+ n \frac{p(w^v_k)p(w^m_i)}{p(w^v_k,w^m_i)}\right].\\
	\end{aligned}
	\end{equation}
\end{small}
Since $w^m_i$ and $w^v_k$ are positive samples, $w^m_i$ and $w^v_k$ are not independent, then $p(w^v_k,w^m_i)-p(w^v_k)p(w^m_i)>0$. Therefore, we can get the following derivation
\begin{equation}
\begin{aligned}
&\mathcal L_{\text {str}}\\
&\geq\sum_{m=1}^{V}\sum_{v=1,v\neq m}^{V}\mathbb{E}_{w_i^m, w_k^{v}}\log \left[ \frac{p(w^v_k,w^m_i)-p(w^v_k)p(w^m_i)}{p(w^v_k,w^m_i)}\right.\\
&\quad\left.+ n \frac{p(w^v_k)p(w^m_i)}{p(w^v_k,w^m_i)}\right]\\
&\geq\sum_{m=1}^{V}\sum_{v=1,v\neq m}^{V}\mathbb{E}_{w_i^m, w_k^{v}}\log \left[ n \frac{p(w^v_k)p(w^m_i)}{p(w^v_k,w^m_i)}\right]\\
&= \sum_{m=1}^{V}\sum_{v=1,v\neq m}^{V}\left[\log (n)-I(w^v_k,w^m_i)\right],\\
\end{aligned}
\end{equation}
where $I(w^v_k,w^m_i)$ represents the mutual information between $w^v_k$ and $w^m_i$.
Therefore, we can get $-\mathcal L_{\text {str}}\leq \sum_{m=1}^{V}\sum_{v=1,v\neq m}^{V}\left[I(w^v_k,w^m_i)-log\left(n\right)\right]$, and minimizing $\mathcal L_{\text {str}}$ is equivalent to maximizing the mutual information of all positive pairs. In fact,  structural-level CL maximizes the mutual information of the subspace structures in any cross views, which captures the nonlinear statistical dependencies among the subspace structures in any cross views, and thus, can serve as a measure of true dependence for exploring the real and reliable structural information of representing accurate categories.\par
\subsection*{\rm {\textbf{Structural-level CL and Probabilistic Intra- and Inter- scatter}}}
The relationship between structural-level CL and probabilistic intra- and inter- scatter are analyzed from the theoretical aspect. When $W^m, m=1,\dots,V$ are fixed, $\mathcal{L}_{str}=\sum_{m=1}^{V} ||Y^m-Y^mW^m||^2_F$.
With simple algebraic computation, we have
\begin{equation}
\begin{aligned}
||Y^m-Y^mW^m||^2_F=Y^m(I-W^m)(I-W^{m})^TY^{m T}.
\end{aligned}
\end{equation}
On the other hand, if we do a normalization of $w^m_k, \forall k$, i.e., $w^m_k = \frac{w^m_k}{||w^m_k||}, \forall k$, we have the following equation:
\begin{equation}
\begin{aligned}
&\sum_{i=1}^{n} \left[(I-W^m)(I-W^{m})^T\right]_{i,k}\\
=&\sum_{i=1}^{n} I_{i,k}-W^m_{i,k}-W^m_{i,k}+(W^mW^{m T})_{i,k}\\
=&1-\sum_{i=1}^{n} W^m_{i,k}-\sum_{i=1}^{n} W^m_{k,i}+\sum_{i=1}^{n}\sum_{j=1}^{n}W^m_{i,j}W^m_{k,j}\\
=&1-1-\sum_{i=1}^{n} W^m_{k,i}+\sum_{i=1}^{n}\sum_{j=1}^{n}W^m_{i,j}W^m_{k,j}\\
=&0.\\
\end{aligned}
\end{equation}
Therefore, the following equation can be obtained:
\begin{equation}
\begin{aligned}
\mathcal{L}_{\text{str}}=&\sum_{m=1}^{V}Y^m(I-W^m)(I-W^{m})^TY^{m T}\\
=&\sum_{m=1}^{V}\sum_{i=1}^{n}||w^m_i||\sum_{k=1}^{n}||y^m_i-y^m_k||^2S^m_{i,k}
\end{aligned}
\end{equation}
where $S^m_{i,k}=(W^m+W^{m T}-W^mW^{m T})_{i,k}$.\par 
Therefore, $S^m_{i,k}$ represents the probability that $y^m_i$ and $y^m_k$ belong to intra-class samples when $S^m_{i,k}\geq 0$, otherwise $|S^m_{i,k}|$ represents the probability that they belong to inter-class samples. In particular, $||w^m_i||$ indicates the reliability of $S^m_{i,k}, \forall k$ calculated by the previous generated subspace structural information. Therefore, structural-level CL actually minimizes the probabilistic intra-scatter and maximizes probabilistic inter-scatter obtained from the potential subspace structures in all views, so as to extract  discriminative features more effectively.\par 
\subsection{Optimization Strategy}
(1) When $P_m, m=1,...,m$ are fixed, the optimization problem (\ref{MFEDCH}) becomes
\begin{equation}
\begin{aligned}
\min_{W^1,...,W^V}~&\sum_{m=1}^{V}\sum_{v=1}^{V}\mathbb{E}_{w_i^m, w_i^{v},\left\{w_{k}^{v}\right\}_{i=1}^{n}}\left[-\log \frac{\operatorname{exp}(\operatorname{sim}\left(w_i^m,  w_i^v\right)}{\sum_{k=1}^{n} \operatorname{exp}(\operatorname{sim}\left(w^m_i,  w_k^{v}\right)}\right]\\
&+\sum_{m=1}^{V}\sum_{i=1}^{n}\left(\alpha\left\|P^T\tilde{x}_i^m-P^T\tilde{X}^mw_i^m\right\|_{2}^{2}+\beta\left\|w_i^m\right\|_2^2\right),
\end{aligned}
\end{equation}
where 
\begin{equation}
P=[P_1;...;P_V]\in R^{D\times d},
\end{equation}
\begin{equation}
\tilde{X}^m=[\underbrace{\mathbf{0};...;\mathbf{0}}_{m-1};X^{m};\underbrace{\mathbf{0};...;\mathbf{0}}_{V-m}]\in R^{D\times n},
\end{equation}
$v$-th $\mathbf{0}$ is the zero matrix of the corresponding scale of the data in the $v$-th view, $\tilde{x}^m_i$ is the $i$-th column vector of $\tilde{X}^m$, $D=\sum_{m=1}^{n}D_m$.\par 
When $w_k^v, k=1,...,n, v=1,...,V, v\neq m$ are fixed, the optimization problem for $w_i^m$ becomes
\begin{equation}\label{w}
\begin{aligned}
&
\min_{w^m_i}~\sum_{v=1}^{V}\left[-\log \frac{\operatorname{exp}(\operatorname{sim}\left(w_i^m,  w_i^v\right)}{\sum_{k=1}^{n} \operatorname{exp}(\operatorname{sim}\left(w^m_i,  w_k^{v}\right)}\right]\\
&\qquad+\alpha\left\|P^T\tilde{x}_i^m-P^T\tilde{X}^mw_i^m\right\|_{2}^{2}+\beta\left\|w_i^m\right\|_2^2
\end{aligned}
\end{equation}
The gradient of the loss function with respect to $w_i^m$ is obtained from (\ref{grad2}).\par 
\begin{equation}\label{grad2}
\begin{aligned}
&\sum_{v=1}^{V} \left[-\frac{\sum_{k=1}^{n} \exp \left(\operatorname{sim}\left(w_{i}^{m}, w_{k}^{v}\right)\right)}{\exp \left(\operatorname{sim}\left(w_{i}^{m}, w_{i}^{v}\right)\right)}\right] \cdot\\
&\left\{\exp \left(\operatorname{sim}\left(w_{i}^{m}, w_{i}^{v}\right)\right) \cdot \bigtriangleup^{m v}_{i i} \cdot \sum_{k=1}^{n} \exp \left(\operatorname{sim}\left( w_{i}^{m}, w_{k}^{v}\right)\right)-\right. \\
&\left.\sum_{k=1}^{n}\left[\exp \left(\operatorname{sim}\left(w_{i}^{m},  w_{k}^{v}\right)\right) \cdot \bigtriangleup^{m v}_{i k}\right] \cdot \exp \left(\operatorname{sim}\left(w_{i}^{m}, w_{i}^{v}\right)\right) \right\}\bigg/\\
&{\left[\sum_{k=1}^{n} \exp \left(\operatorname{sim}\left(w_{i}^{m}, w_{k}^{v}\right)\right)\right]^{2}}+2\alpha
\left(\tilde{X}^{m T} PP^T\tilde{X}^{m}w^m_i-\right.\\
&\left.\tilde{X}^{m T} PP^T\tilde{x}^m_i\right),\\
\end{aligned}
\end{equation}
where
\begin{small}
	\begin{equation}
	\begin{aligned}
	&
	\bigtriangleup_{i k}^{m v}=\\
	&\frac{
		\left\{w_k^v \cdot\left\|w_{i}^{m}\right\|\left\|w_{k}^{v}\right\| \sigma_2-\left[\left(w_{i}^{m T}w_{i}^{m}\right)^{-\frac{1}{2}} \cdot w_{i}^{m} \cdot\left\|w_{k}^{v}\right\| \sigma_2\right] w_{i}^{m T} w_{k}^{v}\right\} }{
		\left(\left\|w_{i}^{m}\right\|\left\|w_{k}^{v}\right\| \sigma_2\right)^{2}}.\\
	\end{aligned}
	\end{equation}
\end{small}
(2) When $W^m,m=1,...,V$ are fixed, the optimization problem (\ref{MFEDCH}) becomes
\begin{small}
	\begin{equation}\label{P}
	\begin{aligned}
	\min_{P} ~&\sum_{m=1}^{V}\sum_{v=1}^{V}\mathbb{E}_{\tilde{x}_i^m, \tilde{x}_i^{v},\left\{\tilde{x}_{k}^{v}\right\}_{i=1}^{n}}\left[-\log \frac{\operatorname{exp}(\operatorname{sim}\left(P^T\tilde{x}_i^m,  P^T\tilde{x}_i^v\right)}{\sum_{k=1}^{n} \operatorname{exp}(\operatorname{sim}\left(P^T\tilde{x}^m_i,  P^T\tilde{x}_k^{v}\right)}\right]\\
	&+\alpha\sum_{m=1}^{V}\left\|P^T\tilde{X}^m-P^T\tilde{X}^mW^m\right\|_{F}^{2}.
	\end{aligned}
	\end{equation}
\end{small}

The gradient of the loss function with respect to the projection matrix $P$ is obtained from (\ref{grad1}).\par 
\begin{footnotesize}
	\begin{equation}\label{grad1}
	\begin{aligned}
	&\sum_{m=1}^{V} \sum_{v=1}^{V} \mathbb{E}_{\tilde{x}_{i}^{m}, \tilde{x}_{i}^{v},\left\{\tilde{x}_{k}^{v}\right\}_{i=1}^{n}}\left[-\frac{\sum_{k=1}^{n} \exp \left(\operatorname{sim}\left(P^{T} \tilde{x}_{i}^{m}, P^{T} \tilde{x}_{k}^{v}\right)\right)}{\exp \left(\operatorname{sim}\left(P^{T} \tilde{x}_{i}^{m}, P^{T} \tilde{x}_{i}^{v}\right)\right)}\right] \cdot\\
	&\left\{\exp \left(\operatorname{sim}\left(P^{T} \tilde{x}_{i}^{m}, P^{T} \tilde{x}_{i}^{v}\right)\right) \cdot \nabla^{m v}_{i i} \cdot \sum_{k=1}^{n} \exp \left(\operatorname{sim}\left(P^{T} \tilde{x}_{i}^{m}, P^{T} \tilde{x}_{k}^{v}\right)\right)-\right. \\
	&\left.\sum_{k=1}^{n}\left[\exp \left(\operatorname{sim}\left(P^{T} \tilde{x}_{i}^{m}, P^{T} \tilde{x}_{k}^{v}\right)\right) \cdot \nabla^{m v}_{i k}\right] \cdot \exp \left(\operatorname{sim}\left(P^{T} \tilde{x}_{i}^{m}, P^{T} \tilde{x}_{i}^{v}\right)\right) \right\}\bigg/\\
	&{\left[\sum_{k=1}^{n} \exp \left(\operatorname{sim}\left(P^{T} \tilde{x}_{i}^{m}, P^{T} \tilde{x}_{k}^{v}\right)\right)\right]^{2}}+2\alpha\sum_{m=1}^{V}
	\left(\tilde{X}^{m} \tilde{X}^{m T}+\right.\\
	&\left.\tilde{X}^{m} W^{m} W^{m T} \tilde{X}^{m T}-\tilde{X}^{m} W\tilde{X}^{m T}-\tilde{X}^{m} W^{m T} \tilde{X}^{m T}\right) \cdot P,\\
	\end{aligned}
	\end{equation}
\end{footnotesize}
where
\begin{equation}
\begin{aligned}
&\nabla_{i k}^{m v}= \\
&\left\{\left(\tilde{x}_{i}^{m} \tilde{x}_{k}^{v T}+\tilde{x}_{k}^{v} \tilde{x}_{i}^{m T}\right) P \cdot\left\|P^{T} \tilde{x}_{i}^{m}\right\|\left\|P^{T} \tilde{x}_{k}^{v}\right\| \sigma_1\right. \\
&-\left[\left(\tilde{x}_{i}^{m T} P P^{T} \tilde{x}_{i}^{m}\right)^{-\frac{1}{2}} \cdot \tilde{x}_{i}^{m} \tilde{x}_{i}^{m T} P \cdot\left\|P^{T} \tilde{x}_{k}^{v}\right\| \sigma_1\right. \\
&\left.\left.+\left(\tilde{x}_{k}^{v T} P P^{T} \tilde{x}_{k}^{v}\right)^{-\frac{1}{2}} \cdot \tilde{x}_{k}^{v} \tilde{x}_{k}^{v T} P \cdot\left\|P^{T} \tilde{x}_{i}^{m}\right\| \sigma\right] \cdot \tilde{x}_{i}^{m T} P P^{T} \tilde{x}_{k}^{v}\right\} \bigg/ \\
&\left(\left\|P^{T} \tilde{x}_{i}^{m}\right\|\left\|P^{T} \tilde{x}_{k}^{v}\right\| \sigma\right)^{2}.
\end{aligned}
\end{equation}

The problem (\ref{MFEDCH}) is solved by using the Adam optimizer\cite{adam}. Adam is an advancement on the random gradient descent method and can rapidly yield accurate result. This method calculates the adaptive learning rate of various parameters based on the budget of the first and second moments of the gradient. The parameters $\gamma$, $\beta_1$, $\beta_2$, and $\epsilon$ represent the learning rate, the exponential decay rate of the first- and second-order moment estimation, and the parameter to prevent division by zero in the implementation, respectively. 
The optimization algorithm for problem (\ref{MFEDCH}) is summarized in the Algorithm 1. The convergent condition used in our experiments
is set as $\left|\mathcal L(P_m^t, m=1,..,V)-\mathcal L(P_m^t+1, m=1,...,V)\right|\leq 10^{-3}$.
\begin{algorithm}[!h]\label{Algorithm1}
	\caption{MFEDCH} 
	{\bf Input:} 
	Data matrix: $X^m\in R^{D_m\times n}, m=1,...,V$, $d$, $\lambda, \alpha,\beta, \gamma, \beta_1,\beta_2,\epsilon$.\\
	$t= 0$ (Initialize number of iterations)\\
	$P_m^0$ (Initialize projection matrix)\\
	\hspace*{0.02in} {\bf Output:} 
	Projection matrix $P^m$
	\begin{algorithmic}
		\WHILE{$P^t_m$ not converged} 
		\FOR{$m=1$ to $V$}
		\FOR{$i=1$ to $n$}
		\STATE
		Updata $w^m_i$ using Adam optimizer by (\ref{grad2})
		\ENDFOR
		\ENDFOR\\
		Updata $P^t_m$ using Adam optimizer by (\ref{grad1})\\
		$t=t+1$\\
		\ENDWHILE
		\RETURN $P^t_m, m=1,...,V$
	\end{algorithmic}
\end{algorithm}

\section{Experiments}
\subsection{Experiments setups}
\begin{table}[!ht]      
	\centering
	\caption{Experimental details about datasets.}
	\label{T01}
	\resizebox{1\textwidth}{!}{
		\begin{tabular}[t]{c c c c c}
			\hline
			Type of experiments&Dataset & View &No. of classes &No. of training samples\\
			\hline
			\multirow{4}{*}{Comparison experiments}&
			Yale&GS, LBP&15&$M$=4,6,8\\
			&ORL&GS, LBP&40&$M$=4,6,8\\
			&Coil-20&GS, LBP&20&$M$=4,6,8\\
			&MF&FAC, FOU, PIX&10&$M$=6,8,10\\
			\hline
			Combination experiments&
			Core1k&GS, ROT, FLI, NOI&10&$M$=20\\
			\hline
		\end{tabular}} 
	\end{table}	
	We demonstrate the strong performance of our framework on classification tasks to assess the effectiveness of our proposed MFEDCH.
	Numerical experiments were set up in two parts, including comparison experiments and combination experiments.
	The comparison experiments are used to demonstrate the performance of MFEDCH over the traditional multi-view feature extraction. The combination experiments are used to demonstrate that structural-level contrastive loss is introduced into other sample-level methods still work well. The $k$-nearest neighbor classifier ($k$= 1) are used in the experiments. Moreover, we randomly select $M$ samples per class for training, whereas the remaining data are used for testing.  The details of each dataset are shown in Table \ref{T01}. All processes are repeated five times, and the final evaluation criteria constitute the classification accuracy of low-dimensional features obtained by \uppercase\expandafter{\romannumeral1} and \uppercase\expandafter{\romannumeral2}. The experiments are implemented using MATLAB R2018a on a computer with an Intel Core i5-9400 2.90 GHz CPU and Windows 10 operating system.\par 
	\uppercase\expandafter{\romannumeral1}. low-dimensional features for each view: 
	\begin{equation}\label{1}
	Y_{m}=P_{m}^{\mathrm{T}} X_{m}, m=1, \ldots, V
	\end{equation}
	
	\uppercase\expandafter{\romannumeral2}. low-dimensional features by fusion strategy:
	\begin{equation}\label{2}
	Y=P_{1}^{\mathrm{T}} X_{1}+\cdots+P_{V}^{\mathrm{T}} X_{V}
	\end{equation}
	\subsection{Experiment Results}
	\subsection*{\bf{Comparison Experiments}}
	\begin{table*}
		\caption{Experimental results of Yale dataset (maximum classification accuracy $\pm$ standard deviations \%).}
		\label{T1}
		\resizebox{\textwidth}{!}{
			\begin{tabular}{c c c c c c c }
				\toprule [2pt]
				View& LPCCA & ALPCCA & GDMCCA & SLCR & KMSA-PCA & MFEDCH\\
				\hline
				\hline
				\multicolumn{7}{c}{\textbf{Train-4}}\\
				\hline
				\textbf{GS}  
				&$66.19\pm1.99$&$63.24\pm3.13$&$73.52\pm2.37$&\bm{$81.52\pm1.86$}&$73.33\pm0.67$&$80.00\pm2.02$\\
				\textbf{LBP}
				&$55.67\pm9.05$&$65.52\pm5.87$&$69.52\pm6.70$&$81.52\pm6.75$&$80.57\pm7.30$&\bm{$88.95\pm3.41$}\\
				\textbf{Mean}	
				&$61.43\pm5.52$&$64.38\pm4.50$&$71.52\pm4.54$&$81.52\pm4.31$&$76.95\pm3.99$&\bm{$84.48\pm2.72$}\\
				\textbf{\uppercase\expandafter{\romannumeral2}} 
				&$64.38\pm6.79$&$69.48\pm5.87$&$74.48\pm5.28$&$76.76\pm6.82$&$75.05\pm4.28$&\bm{$79.62\pm3.60$}\\
				\hline
				\hline
				
				\multicolumn{7}{c}{\textbf{Train-6}}\\
				\hline
				\textbf{GS}  
				&$63.33\pm2.40$&$64.67\pm9.09$&$79.20\pm3.96$&\bm{$80.27\pm4.75$}&$73.07\pm4.36$&$79.20\pm5.04$\\
				\textbf{LBP}
				&$64.00\pm6.53$&$67.07\pm7.02$&$68.00\pm1.89$&$89.07\pm2.19$&$83.20\pm3.84$&\bm{$93.07\pm2.19$}\\		
				\textbf{Mean}	
				&$63.67\pm4.47$&$65.87\pm8.06$&$73.60\pm2.93$&$84.67\pm2.21$&$78.13\pm4.10$&\bm{$86.13\pm3.62$}\\
				\textbf{\uppercase\expandafter{\romannumeral2}} 
				&$67.60\pm4.91$&$76.27\pm3.67$&$79.47\pm4.28$&\bm{$80.80\pm3.48$}&$74.93\pm4.36$&$78.93\pm3.18$\\
				\hline
				\hline
				
				\multicolumn{7}{c}{\textbf{Train-8}}\\
				\hline
				\textbf{GS}  
				&$66.22\pm7.27$&$76.00\pm3.65$&$80.44\pm5.75$&$88.00\pm7.30$&$80.00\pm7.86$&\bm{$88.52\pm4.35$}\\
				\textbf{LBP}
				&$79.56\pm2.90$&$76.44\pm9.07$&$75.56\pm4.97$&$91.56\pm5.31$&$86.67\pm7.03$&\bm{$92.36\pm3.96$}\\		
				\textbf{Mean}	
				&$72.89\pm5.09$&$76.22\pm6.36$&$78.00\pm5.36$&$89.78\pm6.31$&$83.33\pm7.45$&\bm{$90.44\pm4.16$}\\
				\textbf{\uppercase\expandafter{\romannumeral2}}
				&$76.89\pm4.33$&$82.67\pm3.65$&$85.33\pm4.33$&$90.33\pm2.72$&$80.44\pm6.36$&\bm{$90.67\pm3.65$}\\
				\bottomrule[2pt]
			\end{tabular}}
		\end{table*}	
		
		\begin{table*}
			\caption{Experimental results of ORL dataset (maximum classification accuracy ± standard deviations \%).}
			\label{T2}
			\resizebox{\textwidth}{!}{
				\begin{tabular}{c c c c c c c }
					\toprule [2pt]
					View& LPCCA & ALPCCA & GDMCCA & SLCR & KMSA-PCA & MFEDCH\\
					\hline
					\hline
					\multicolumn{7}{c}{\textbf{Train-4}}\\
					\hline
					\textbf{GS}  
					&$84.08\pm1.75$&$82.75\pm1.90$&$82.17\pm1.90$&$88.17\pm1.78$&$81.00\pm2.68$&\bm{$90.33\pm2.01$}\\
					\textbf{LBP}
					&$71.29\pm0.73$&$72.83\pm2.66$&$69.00\pm2.03$&$76.75\pm1.46$&$76.58\pm0.85$&\bm{$79.83\pm1.81$}\\
					\textbf{Mean}	
					&$77.69\pm1.24$&$77.79\pm2.28$&$75.58\pm1.97$&$82.46\pm1.62$&$78.79\pm1.77$&\bm{$85.08\pm1.91$}\\
					\textbf{\uppercase\expandafter{\romannumeral2}}
					&$75.46\pm2.20$&$81.58\pm3.38$&$79.17\pm1.98$&\bm{$83.92\pm1.92$}&$83.67\pm2.85$&$83.75\pm3.97$\\
					\hline
					\hline
					
					\multicolumn{7}{c}{\textbf{Train-6}}\\
					\hline
					\textbf{GS}  
					&$91.75\pm1.79$&$91.00\pm1.30$&$89.38\pm1.17$&$92.13\pm2.01$&$84.25\pm1.84$&\bm{$95.50\pm1.03$}\\
					\textbf{LBP}
					&$82.25\pm2.01$&$85.13\pm1.56$&$83.50\pm1.69$&$86.50\pm2.64$&$82.37\pm0.81$&\bm{$88.38\pm1.51$}\\		
					\textbf{Mean}	
					&$87.00\pm1.90$&$88.06\pm1.43$&$86.44\pm1.43$&$89.31\pm2.33$&$83.31\pm1.33$&\bm{$91.94\pm1.27$}\\
					\textbf{\uppercase\expandafter{\romannumeral2}}
					&$87.37\pm1.54$&$90.50\pm1.35$&$89.00\pm1.69$&$91.38\pm1.43$&$87.50\pm1.40$&\bm{$91.75\pm2.04$}\\
					\hline
					\hline
					
					\multicolumn{7}{c}{\textbf{Train-8}}\\
					\hline
					\textbf{GS}  
					&$94.25\pm4.01$&$93.75\pm3.19$&$91.75\pm4.20$&$94.00\pm2.56$&$89.50\pm4.01$&\bm{$97.50\pm1.05$}\\
					\textbf{LBP}
					&$85.50\pm4.47$&$88.75\pm2.93$&$88.00\pm3.71$&$89.25\pm3.26$&$88.00\pm4.56$&\bm{$90.50\pm2.85$}\\		
					\textbf{Mean}	
					&$89.88\pm4.24$&$91.25\pm3.06$&$89.88\pm3.96$&$91.63\pm2.91$&$88.75\pm4.29$&\bm{$94.13\pm1.95$}\\
					\textbf{\uppercase\expandafter{\romannumeral2}} 
					&$90.12\pm2.81$&$92.25\pm1.63$&$91.25\pm3.19$&$92.50\pm3.64$&$91.75\pm2.59$&\bm{$93.25\pm2.74$}\\
					\bottomrule[2pt]
				\end{tabular}}
			\end{table*}	
					To demonstrate the effectiveness of MFEDCH for multi-view feature extraction, we compare it on Yale, ORL, Coil-20, and MF datasets with the traditional multi-view feature extraction methods of LPCCA\cite{lpcca}, ALPCCA\cite{alpcca}, GDMCCA\cite{gmcca}, SLCR\cite{slcr}, and KMSA-PCA\cite{kmsa}.\par
					The more appropriate default parameters for testing machine learning problems in Adam optimizer comprise
					$\gamma = 0.001$, $\beta_1 = 0.9$, $\beta_2 = 0.999$, and $\epsilon = 10^{-8}$.  Furthermore, the ranges of $\sigma_1$ and $\sigma_2$ are set as $\{0.01, 0.1, 1, 10, 100, 1000\}$, and we fixed $\lambda=1$, $\alpha=1$, $\beta=1$.\par
					 The best results are highlighted in boldface. ``\textbf{Train-}\bm{$M$}" represent the classification accuracy of $M$ training samples for per class, and ``\textbf{Mean}" represent the average classification accuracy of all views of the low-dimensional features obtained  by \uppercase\expandafter{\romannumeral1}.  ``\textbf{\uppercase\expandafter{\romannumeral2}}" represent the classification accuracy of the low-dimensional features obtained  by \uppercase\expandafter{\romannumeral2}. 
					\par 
					On the Yale dataset:\par 
					
					As can be observed from Table \ref{T1}, the maximum average classification accuracy (i.e., \textbf{Mean}) are always achieved by our proposed MFEDCH. In most cases, the other classification accuracy of MFEDCH are higher than that of the comparison methods. Only the classification accuracy of \textbf{GS} in \textbf{Train-4,6} and \textbf{\uppercase\expandafter{\romannumeral2}} in \textbf{Train-6} are the highest after using SLCR feature extraction. \par
					
					On the ORL dataset:\par
					As can be observed from Table \ref{T2},  the maximum average classification accuracy (i.e., \textbf{Mean}) are always achieved by our proposed MFEDCH. In most cases, the other classification accuracy of MFEDCH are higher than that of the comparison methods. Only the classification accuracy of \textbf{\uppercase\expandafter{\romannumeral2}} in \textbf{Train-4} are the highest after using SLCR feature extraction. \par

						\section{Conclusion}
						In this study, we have proposed a multi-view feature extraction method based on dual contrastive head (MFEDCH), which introduces structural-level contrastive loss into the sample-level CL-based method CMC. In MFEDCH, structural-level CL push the potential subspace structures consistent in any cross views, which assists sample-level CL to extract discriminative features more effectively. In addition, strucural-level CL can also be combined with other sample-level CL-based methods, such as SimCLR, InvP, MoCHi, PCL, HCSC, BYOL, SimSiam, etc., which provides a new framework for future CL. In theory, it has been proven that the relationships between structural-level CL and mutual information and probabilistic intra- and inter- scatter, which provides the theoretical support for the excellent performance. The numerical experiments
						demonstrate that the proposed method offers a strong
						advantage for multi-view feature extraction.\par 
						However, our framework contains many parameters, so it is difficult and
						costly to adjust all parameters to obtain the optimal projections in the context of big data. We will continue to work on the research of feature extraction algorithms based on CL in the future.
						
						\bibliography{bibfile}

\begin{thebibliography}{10}
\expandafter\ifx\csname url\endcsname\relax
  \def\url#1{\texttt{#1}}\fi
\expandafter\ifx\csname urlprefix\endcsname\relax\def\urlprefix{URL }\fi
\expandafter\ifx\csname href\endcsname\relax
  \def\href#1#2{#2} \def\path#1{#1}\fi

\bibitem{c1}
A.~Huang, Z.~Wang, Y.~Zheng, T.~Zhao, C.~Lin, Embedding regularizer learning
  for multi-view semi-supervised classification, {IEEE} Trans. Image Process.
  30 (2021) 6997--7011.
\newblock \href {https://doi.org/10.1109/TIP.2021.3101917}
  {\path{doi:10.1109/TIP.2021.3101917}}.

\bibitem{cluster1}
A.~Khan, P.~Maji, Multi-manifold optimization for multi-view subspace
  clustering, {IEEE} Trans. Neural Networks Learn. Syst. 33~(8) (2022)
  3895--3907.
\newblock \href {https://doi.org/10.1109/TNNLS.2021.3054789}
  {\path{doi:10.1109/TNNLS.2021.3054789}}.

\bibitem{cluster2}
C.~Zhang, H.~Fu, Q.~Hu, X.~Cao, Y.~Xie, D.~Tao, D.~Xu, Generalized latent
  multi-view subspace clustering, {IEEE} Trans. Pattern Anal. Mach. Intell.
  42~(1) (2020) 86--99.
\newblock \href {https://doi.org/10.1109/TPAMI.2018.2877660}
  {\path{doi:10.1109/TPAMI.2018.2877660}}.

\bibitem{e1}
L.~Zhao, T.~Yang, J.~Zhang, Z.~Chen, Y.~Yang, Z.~J. Wang, Co-learning
  non-negative correlated and uncorrelated features for multi-view data, {IEEE}
  Trans. Neural Networks Learn. Syst. 32~(4) (2021) 1486--1496.
\newblock \href {https://doi.org/10.1109/TNNLS.2020.2984810}
  {\path{doi:10.1109/TNNLS.2020.2984810}}.

\bibitem{s1}
C.~Tang, X.~Zheng, X.~Liu, W.~Zhang, J.~Zhang, J.~Xiong, L.~Wang, Cross-view
  locality preserved diversity and consensus learning for multi-view
  unsupervised feature selection, {IEEE} Trans. Knowl. Data Eng. 34~(10) (2022)
  4705--4716.
\newblock \href {https://doi.org/10.1109/TKDE.2020.3048678}
  {\path{doi:10.1109/TKDE.2020.3048678}}.

\bibitem{ex1}
C.~Zhang, H.~Fu, Q.~Hu, P.~Zhu, X.~Cao, Flexible multi-view dimensionality
  co-reduction, {IEEE} Trans. Image Process. 26~(2) (2017) 648--659.
\newblock \href {https://doi.org/10.1109/TIP.2016.2627806}
  {\path{doi:10.1109/TIP.2016.2627806}}.

\bibitem{ex2}
W.~Zhuge, F.~Nie, C.~Hou, D.~Yi, Unsupervised single and multiple views feature
  extraction with structured graph, {IEEE} Trans. Knowl. Data Eng. 29~(10)
  (2017) 2347--2359.
\newblock \href {https://doi.org/10.1109/TKDE.2017.2725263}
  {\path{doi:10.1109/TKDE.2017.2725263}}.

\bibitem{ex3}
J.~Zhang, L.~Liu, L.~Zhen, L.~Jing, A unified robust framework for multi-view
  feature extraction with l2, 1-norm constraint, Neural Networks 128 (2020)
  126--141.
\newblock \href {https://doi.org/10.1016/j.neunet.2020.04.024}
  {\path{doi:10.1016/j.neunet.2020.04.024}}.

\bibitem{cl1}
E.~Pan, Z.~Kang, Multi-view contrastive graph clustering, in: M.~Ranzato,
  A.~Beygelzimer, Y.~N. Dauphin, P.~Liang, J.~W. Vaughan (Eds.), Proc. NeurIPS,
  2021, pp. 2148--2159.

\bibitem{cl2}
T.~Wang, P.~Isola, Understanding contrastive representation learning through
  alignment and uniformity on the hypersphere, in: Proc. ICML, Vol. 119, 2020,
  pp. 9929--9939.

\bibitem{cl3}
P.~Khosla, P.~Teterwak, C.~Wang, A.~Sarna, Y.~Tian, P.~Isola, A.~Maschinot,
  C.~Liu, D.~Krishnan, Supervised contrastive learning, in: Proc. NeurIPS,
  2020.

\bibitem{cl4}
J.~Z. HaoChen, C.~Wei, A.~Gaidon, T.~Ma, Provable guarantees for
  self-supervised deep learning with spectral contrastive loss, in: Proc.
  NeurIPS, 2021, pp. 5000--5011.

\bibitem{cl5}
N.~Saunshi, O.~Plevrakis, S.~Arora, M.~Khodak, H.~Khandeparkar, A theoretical
  analysis of contrastive unsupervised representation learning, in: Proc. ICML,
  Vol.~97, 2019, pp. 5628--5637.

\bibitem{cpc}
A.~van~den Oord, Y.~Li, O.~Vinyals, Representation learning with contrastive
  predictive coding, arXiv preprint (2018).
\newblock \href {http://arxiv.org/abs/1807.03748} {\path{arXiv:1807.03748}}.

\bibitem{cmc}
Y.~Tian, D.~Krishnan, P.~Isola, Contrastive multiview coding, in: {Proc. ECCV},
  Vol. 12356, 2020, pp. 776--794.

\bibitem{simclr}
T.~Chen, S.~Kornblith, M.~Norouzi, G.~E. Hinton, A simple framework for
  contrastive learning of visual representations, in: {Proc. ICML}, Vol. 119,
  2020, pp. 1597--1607.

\bibitem{npid}
Z.~Wu, Y.~Xiong, S.~X. Yu, D.~Lin, Unsupervised feature learning via
  non-parametric instance discrimination, in: {Proc. CVPR}, 2018, pp.
  3733--3742.
\newblock \href {https://doi.org/10.1109/CVPR.2018.00393}
  {\path{doi:10.1109/CVPR.2018.00393}}.

\bibitem{pirl}
I.~Misra, L.~van~der Maaten, Self-supervised learning of pretext-invariant
  representations, in: Proc. CVPR, 2020, pp. 6706--6716.
\newblock \href {https://doi.org/10.1109/CVPR42600.2020.00674}
  {\path{doi:10.1109/CVPR42600.2020.00674}}.

\bibitem{moco}
K.~He, H.~Fan, Y.~Wu, S.~Xie, R.~B. Girshick, Momentum contrast for
  unsupervised visual representation learning, in: Proc. CVPR, 2020, pp.
  9726--9735.
\newblock \href {https://doi.org/10.1109/CVPR42600.2020.00975}
  {\path{doi:10.1109/CVPR42600.2020.00975}}.

\bibitem{invp}
F.~Wang, H.~Liu, D.~Guo, F.~Sun, Unsupervised representation learning by
  invariance propagation, in: Proc. NeurIPS, 2020.

\bibitem{mochi}
Y.~Kalantidis, M.~B. Sariyildiz, N.~Pion, P.~Weinzaepfel, D.~Larlus, Hard
  negative mixing for contrastive learning, in: Proc. NeurIPS, 2020.

\bibitem{pcl}
J.~Li, P.~Zhou, C.~Xiong, S.~C.~H. Hoi, Prototypical contrastive learning of
  unsupervised representations, in: {Proc. ICLR}, OpenReview.net, 2021.

\bibitem{hcsc}
Y.~Guo, M.~Xu, J.~Li, B.~Ni, X.~Zhu, Z.~Sun, Y.~Xu, {HCSC:} hierarchical
  contrastive selective coding, in: Proc. CVPR, 2022, pp. 9696--9705.
\newblock \href {https://doi.org/10.1109/CVPR52688.2022.00948}
  {\path{doi:10.1109/CVPR52688.2022.00948}}.

\bibitem{cca}
D.~R. Hardoon, S.~Szedm{\'{a}}k, J.~Shawe{-}Taylor, Canonical correlation
  analysis: An overview with application to learning methods, Neural Comput.
  16~(12) (2004) 2639--2664.
\newblock \href {https://doi.org/10.1162/0899766042321814}
  {\path{doi:10.1162/0899766042321814}}.

\bibitem{lpcca}
T.~Sun, S.~Chen, Locality preserving {CCA} with applications to data
  visualization and pose estimation, Image Vis. Comput. 25~(5) (2007) 531--543.
\newblock \href {https://doi.org/10.1016/j.imavis.2006.04.014}
  {\path{doi:10.1016/j.imavis.2006.04.014}}.

\bibitem{alpcca}
F.~Wang, D.~Zhang, A new locality-preserving canonical correlation analysis
  algorithm for multi-view dimensionality reduction, Neural Process. Lett.
  37~(2) (2013) 135--146.
\newblock \href {https://doi.org/10.1007/s11063-012-9238-9}
  {\path{doi:10.1007/s11063-012-9238-9}}.

\bibitem{cscca}
C.~Zu, D.~Zhang, Canonical sparse cross-view correlation analysis,
  Neurocomputing 191 (2016) 263--272.
\newblock \href {https://doi.org/10.1016/j.neucom.2016.01.053}
  {\path{doi:10.1016/j.neucom.2016.01.053}}.

\bibitem{wcscca}
C.~Zhu, R.~Zhou, C.~Zu, Weight-based canonical sparse cross-view correlation
  analysis, Pattern Anal. Appl. 22~(2) (2019) 457--476.
\newblock \href {https://doi.org/10.1007/s10044-017-0644-5}
  {\path{doi:10.1007/s10044-017-0644-5}}.

\bibitem{colpp}
X.~Zhao, X.~Wang, H.~Wang, Multi-view dimensionality reduction via subspace
  structure agreement, Multim. Tools Appl. 76~(16) (2017) 17437--17460.
\newblock \href {https://doi.org/10.1007/s11042-016-3943-8}
  {\path{doi:10.1007/s11042-016-3943-8}}.

\bibitem{mcca}
J.~Rupnik, J.~Shawe-Taylor, Multi-view canonical correlation analysis, Taylor
  (2010) 1--4.

\bibitem{mvpls}
G.~Cao, A.~Iosifidis, K.~Chen, M.~Gabbouj, Generalized multi-view embedding for
  visual recognition and cross-modal retrieval, {IEEE} Trans. Cybern. 48~(9)
  (2018) 2542--2555.
\newblock \href {https://doi.org/10.1109/TCYB.2017.2742705}
  {\path{doi:10.1109/TCYB.2017.2742705}}.

\bibitem{le}
M.~Belkin, P.~Niyogi, Laplacian eigenmaps and spectral techniques for embedding
  and clustering, in: T.~G. Dietterich, S.~Becker, Z.~Ghahramani (Eds.), Proc.
  NIPS, 2001, pp. 585--591.

\bibitem{mse}
T.~Xia, D.~Tao, T.~Mei, Y.~Zhang, Multiview spectral embedding, {IEEE} Trans.
  Syst. Man Cybern. Part {B} 40~(6) (2010) 1438--1446.
\newblock \href {https://doi.org/10.1109/TSMCB.2009.2039566}
  {\path{doi:10.1109/TSMCB.2009.2039566}}.

\bibitem{mrpe}
H.~Wang, L.~Feng, A.~Kong, B.~Jin, Multi-view reconstructive preserving
  embedding for dimension reduction, Soft Comput. 24~(10) (2020) 7769--7780.
\newblock \href {https://doi.org/10.1007/s00500-019-04395-4}
  {\path{doi:10.1007/s00500-019-04395-4}}.

\bibitem{cmsre}
H.~Wang, J.~Peng, X.~Fu, Co-regularized multi-view sparse reconstruction
  embedding for dimension reduction, Neurocomputing 347 (2019) 191--199.
\newblock \href {https://doi.org/10.1016/j.neucom.2019.03.080}
  {\path{doi:10.1016/j.neucom.2019.03.080}}.

\bibitem{spmcca}
L.~Gao, L.~Qi, L.~Guan, Sparsity preserving multiple canonical correlation
  analysis with visual emotion recognition to multi-feature fusion, in: Proc.
  ICIP, 2015, pp. 2710--2714.
\newblock \href {https://doi.org/10.1109/ICIP.2015.7351295}
  {\path{doi:10.1109/ICIP.2015.7351295}}.

\bibitem{gmcca}
J.~Chen, G.~Wang, G.~B. Giannakis, Graph multiview canonical correlation
  analysis, {IEEE} Trans. Signal Process. 67~(11) (2019) 2826--2838.
\newblock \href {https://doi.org/10.1109/TSP.2019.2910475}
  {\path{doi:10.1109/TSP.2019.2910475}}.

\bibitem{kmsa}
H.~Wang, Y.~Wang, Z.~Zhang, X.~Fu, L.~Zhuo, M.~Xu, M.~Wang, Kernelized
  multiview subspace analysis by self-weighted learning, {IEEE} Trans. Multim.
  23 (2021) 3828--3840.
\newblock \href {https://doi.org/10.1109/TMM.2020.3032023}
  {\path{doi:10.1109/TMM.2020.3032023}}.

\bibitem{slcr}
J.~Zhang, H.~Zhang, W.~Qiang, N.~Deng, L.~Jing, Locality cross-view regression
  for feature extraction, Eng. Appl. Artif. Intell. 105 (2021) 104414.
\newblock \href {https://doi.org/10.1016/j.engappai.2021.104414}
  {\path{doi:10.1016/j.engappai.2021.104414}}.

\bibitem{adam}
D.~P. Kingma, J.~Ba, Adam: A method for stochastic optimization, in: {Proc.
  ICLR} (Poster), 2015.

\bibitem{byol}
J.~Grill, F.~Strub, F.~Altch{\'{e}}, C.~Tallec, P.~H. Richemond,
  E.~Buchatskaya, C.~Doersch, B.~{\'{A}}. Pires, Z.~Guo, M.~G. Azar, B.~Piot,
  K.~Kavukcuoglu, R.~Munos, M.~Valko, Bootstrap your own latent - {A} new
  approach to self-supervised learning, in: Proc. NeurIPS, 2020.

\bibitem{simsiam}
X.~Chen, K.~He, Exploring simple siamese representation learning, in: Proc.
  CVPR, 2021, pp. 15750--15758.
\newblock \href {https://doi.org/10.1109/CVPR46437.2021.01549}
  {\path{doi:10.1109/CVPR46437.2021.01549}}.

\end{thebibliography}
					\end{document}